\documentclass[11pt,a4paper]{article}
\usepackage[hyperref]{emnlp2020}
\usepackage{times}
\usepackage{latexsym}

\usepackage{microtype}
\usepackage{graphicx}
\usepackage{amsmath}
\usepackage{amssymb}
\usepackage{booktabs}
\usepackage{subcaption}

\usepackage{tikz}
\pgfdeclarelayer{nodelayer}
\pgfdeclarelayer{edgelayer}
\pgfsetlayers{edgelayer,nodelayer,main}

\tikzstyle{var}=[fill=none, draw=none, outer sep=2]
\tikzstyle{var block}=[fill=white, draw=black, shape=rectangle, outer sep=2]
\tikzstyle{dot}=[fill=black, draw=black, shape=circle, scale=0.3]
\tikzstyle{blurb}=[fill=none, draw=none, align=left]

\tikzstyle{arrow}=[->]
\tikzstyle{callout}=[-, dash pattern=on 2pt off 2pt, draw={rgb,255: red,128; green,128; blue,128}]

\aclfinalcopy 

\title{Incorporating Behavioral Hypotheses for Query Generation}

\author{Ruey-Cheng Chen\\
  SEEK Ltd \\
  \texttt{rcchen@seek.com.au} \\\And
  Chia-Jung Lee \\
  Microsoft \\
  \texttt{cjlee@microsoft.com} \\}

\date{}

\begin{document}
\maketitle

\begin{abstract}
    
    Generative neural networks have been shown effective on query suggestion. 
    Commonly posed as a conditional generation problem, the task aims to leverage earlier inputs from users in a search session to predict queries that they will likely issue at a later time.
    User inputs come in various forms such as querying and clicking, each of which can imply different semantic signals channeled through the corresponding behavioral patterns. 
    This paper induces these behavioral biases as hypotheses for query generation, where a generic encoder-decoder Transformer framework is presented to aggregate arbitrary hypotheses of choice.
    Our experimental results show that the proposed approach leads to significant improvements on top-$k$ word error rate and Bert F1 Score compared to a recent BART model.

\end{abstract}

\section{Introduction}

Query suggestion is key to the usability of a search engine in the way it helps users formulating more effective queries or exploring related search needs. 
Prior work tackles this problem by employing primarily two strategies.
The first one is based on a discriminative characterization, with candidate queries drawn from production logs ranked to align with what users may most likely issue next. 
Although effective \cite{ahmad2018multi}, this strategy is inherently restricted by what is available in the logs, which in turn can penalize tail queries \cite{dehghani2017learning}. 
In this work, we pursue the second strategy where query suggestion is cast as a natural language generation problem, aiming at producing effective continuations of the user's intent by using generative modeling.

\begin{figure}[t]
    \centering
    \includegraphics[width=0.8\columnwidth]{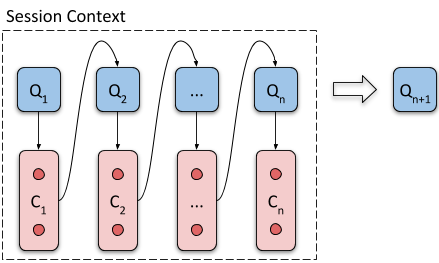}
    \caption{An example search session where a user issues queries and optionally performs clicking at timestamps $1$ to $n$. At time $n$+$1$, the user issues $q_{n+1}$ following the previous search context of length $n$.}
    \label{f:mrf}
\end{figure}

For query generation, prior research has focused mostly on extending standard Seq2Seq models where the input is a concatenation of earlier queries a user has submitted in a session  \cite{sordoni2015hierarchical, dehghani2017learning}.
However, literature often leaves out the influence of clickthrough actions (i.e., red blocks in Figure~\ref{f:mrf}), which we argue should be taken into account in the generative process as they could be surrogates of the user's implicit search intent \cite{yin2016ranking}.
Users may exhibit diverse behaviors such as consecutively issuing queries without further engagement, or following up a single query with extensive clickthrough actions.
These vastly different patterns are indicative of information pieces that the users find most relevant, which we conjecture can help producing suggestions better aligned to the user needs.

We present an encoder-decoder Transformer model for the generative task that includes these patterns that we called behavioral hypotheses. 
One challenge that arises with Transformers is that they make minimal assumptions about input (i.e., a single string of tokens), making it non-trivial to add multiple hypotheses directly.
To address this issue, we propose a generic approach that leverages tokenwise attention to aggregate multiple behavior hypotheses encoded by a shared Transformer encoder BART \cite{lewis2019bart}. The resulting end-to-end model can capture the underlying user-induced belief while maintaining the same order of complexity as the original BART.
For evaluation, we conduct experiments by sampling over 600K search sessions from a major commercial job search engine in Australia.
With evaluation metrics including word error rate and BertScore \cite{zhang2020bertscore}, we show that the approach outperforms prior competitive baselines and a recent Transformer model BART, suggesting attending to behavioral patterns is crucial to reflect users' intent.

\section{Related Work}

Generative approaches have been studied extensively in machine translation \cite{sutskever2014sequence, bahdanau2014neural}, dialogue systems \cite{wen2015semantically}, and many other related areas \cite{gatt2018survey}. 
The methodology was first applied to the query domain in \citet{sordoni2015hierarchical}.
Query suggestion has traditionally been a web search usability task.
Ranking based approaches that leverage query co-occurrence and discriminative modeling are known to be most effective \cite{ozertem2012learning}, but also likely to suffer from the lack of appropriate candidates for rarely seen queries.
Some recent work sought to characterize the generative nature \cite{sordoni2015hierarchical, dehghani2017learning} of this process.
The hierarchical formulation of sequence-to-sequence model \cite{sordoni2015hierarchical} can effectively capture the query transitions, but does not offer a mechanism to incorporate implicit user signals \cite{wu2018query}.
Our approach combines heterogeneous behavioral hypotheses by leveraging large-scale encoders and cross-structure attentions.
Apart from similar attempts regarding encoding multiple sentences \cite{dai2019transformer, zhao2020transformer}, our work in a generative setting tackles a different problem of decoding over a meshed representation originated from multiple sources.

\section{Approach}
\label{sec:approach}

Let $\mathcal{Q} = (Q_1, Q_2, \ldots, Q_n)$ represent a sequence of queries submitted by a user in a consecutive fashion, where each $Q_i$ comprises a sequence of terms.
Each $Q_i$ can lead to a succession of follow-up user interactions, among which we are mostly interested in textual matching cues $C_i$ that enticed clicking on some underlying documents.
The full list of such matching cues are denoted as $\mathcal{C} = (C_1, C_2, \ldots, C_n)$, in which each $C_i$ is a set of text excerpts $t_{i1}, \ldots, t_{im}$, such as document title or metadata, displayed in response to $Q_i$. 
Given such a search context $(\mathcal{Q}, \mathcal{C})$, the query generation task aims to create a candidate query $Q_{n+1}$ that the user is most likely to follow up with. The overall process is depicted in Figure~\ref{f:mrf}.

\paragraph{Behavioral Hypotheses.}
We conjecture that, when making a new query, the assumed user takes inspiration from his/her preceding search context, following some \emph{behavioral hypotheses} formed by preceding queries or matching cues.
In this paper we seek to characterize this influence to formulated queries as follows:
\begin{equation}
\begin{split}
    K_1 &= (Q_1, Q_2, \ldots, Q_n) \\
    K_2 &= \bigoplus\nolimits_{i<n} \left\{ (t)_{t \in C_i} \right\} \oplus (Q_n) \\
    K_3 &= \bigoplus\nolimits_{Q_i: C_i \ne \varnothing} \left\{ (Q_i) \oplus (t)_{t \in C_i} \right\} \oplus (Q_n) \\
    K_4 &= (Q_n) \oplus (t)_{t \in C_n} \\
\end{split}
\end{equation}
Each of these definitions loosely specifies a generative story behind the process: influence may come directly from preceding queries ($K_1$), preceding matching cues ($K_2$),
interacted queries and the respective matching cues ($K_3$)
, or the most recently submitted query and observed cues ($K_4$).

\paragraph{Vanilla Encoder-Decoder Transformer.}

Recent advances in transfer learning has popularized the use of pretrained encoder-decoder Transformer networks, setting state-of-the-art across the board \cite{vaswani2017attention, raffel2019exploring, lewis2019bart}.
Query generation can be cast as a sequence-to-sequence problem and fine-tuned on any of these pretrained models.
A simplistic but typical approach is, on the input side, to concatenate all items in a behavioral hypothesis regardless of their types into one sentence with the separator token inserted in between, and on the output side to simply put in the ground truth to be generated, i.e., $Q_{n+1}$.
All input/output sentences are first tokenized using the same byte-pair encoding, and properly formatted by adding start/end tokens to the sentence beginning and sentence end.
Following this preparation step, the input sentence is encoded into a vector representation by multiple layers of Transformer, and decoded on the other side using a similar stack.

In this paper we use the BART model \cite{lewis2019bart} to implement this encoder-decoder network.
BART leverages specialized pretraining objectives such as text infilling and is known to be performant on text generation problems such as machine translation and summarization.

\paragraph{Meshed Representations.}

One caveat of the above process is that the model is agnostic of the presence of multiple behavioral hypotheses in the input.
To solve this problem, we propose a new approach that derives a \emph{meshed} representation of the input  hypotheses $K_1, \ldots, K_4$ by reusing one single BART instance.
Each hypothesis is first encoded using a shared BART encoder, and a tokenwise attention mechanism is leveraged to learn effective ways to contextualize the individual representations together.
This is essentially combining four input streams in a token-by-token fashion so that each hypothesis can contribute to the aggregate at each token step, in varying degrees as determined by the attention weights.
We expect this change to help surfacing regularities across different hypotheses, in the way regularization or multi-task learning does to ease learning trajectories.
Tokenwise attention is also designed to encourage early correction in the hope that a more robust representation can be formed at the end of the meshed sequence.

The procedure can be described as follows.
Let $[S_i^{(1)}; \ldots; S_i^{(T)}] = \text{BART}_\text{enc}(K_i)$ for all $K_i$, and $T$ is the sequence length.
We have:
\begin{equation}
\begin{split}
    \alpha_i^{(j)} &\propto \exp(W_\text{attn} S_i^{(j)}) \\
    F^{(j)} &= \sum\nolimits_i \alpha_i^{(j)} S_i^{(j)} \\
    O &= \text{BART}_\text{dec}([F^{(1)}; \ldots; F^{(T)}])
\end{split}
\end{equation}
where $W_\text{attn}$ is the attention weight matrix to be learned and $O$ the output.
On the decoder side, the attention mask is set to the union of attention masks from all underpinning hypotheses.
This approach does not require multiple BART instances but may take extra GPU memory, linear to the output batch size, to cache processed representations and additional computation cycles to work through all four behavioral hypotheses.

\section{Experimental Setup}

Our main testbed was a sample of session logs from the SEEK job search engine\footnote{\url{https://www.seek.com.au}}.
This task domain is known for its characteristic query topics, surrounding role titles, skills, and entities such as company names or geo-locations, and distinctively different user behaviors to general web search.
This dataset is preferred over the AOL logs for the availability of clicked document texts, but our approach should be equally applicable to other search domains.

We collected textual queries $Q_i$ and the titles of documents that were clicked on in response to $Q_i$ as $C_i$.
All search sessions were anonymized to ensure that the query and click information cannot be linked back to individual users.
Session boundaries were determined by an inactivity of 30 minutes or more between two consecutive actions.
In each session the latest query was held out as the ground truth.
Training sessions (500K) were initially gathered from a two-week span starting from Oct 1, 2019, and out of the same period a separate split was selected as a dev set (1K); then, the latter two weeks from the same month were sampled to form the test sessions (100K). 
About 15\% of the collected sessions were found to exceed the maximum sequence length of the BART encoder, and were removed from the experiment to avoid inadvertently favoring the proposed approach for that the baseline may only see truncated input.
Standard preprocessing steps were performed to remove noisy queries that occurred 10 times or less across the periods, and singleton sessions that contain only one query.
In our experiments we compare the following approaches:

\paragraph{Seq2Seq+Attn}

A standard sequence-to-sequence model using a two-layer bidirectional GRU \cite{cho2014learning} as the encoder and a uni-directional attentive GRU as the decoder \cite{bahdanau2014neural}. 
Our implementation used 1,000 hidden dimensions and the same byte-pair encodings as other methods.

\paragraph{MPS (Most Popular Suggestion)}
A simple yet effective baseline used in \cite{hasan2011query, sordoni2015hierarchical, dehghani2017learning}, based on co-occurrence frequencies of the last query in the search context and all candidate queries.

\paragraph{BART}
The vanilla BART model \cite{lewis2019bart}.
We took the full concatenated search context as input, and fine-tuned on pretrained weights for BART-Large model, complete with 12 transformer layers in total.

\paragraph{MeshBART}
The proposed meshed variant of BART, configured to have the same model capacity.
It takes multiple input hypotheses and combines them using the proposed tokenwise attentions before entering the decoding phrase.

\vspace{1ex}

\noindent We report word error rate (WER) and Bert F1 scores (BertF1) \cite{popovic2007word, zhang2020bertscore} adapted to the top-$k$ setting, with respect to the reference (ground truth) across the given $k$ hypotheses.
\[ 
\begin{split}
\text{WER@$k$} &= \min_{i = 1, \ldots, k} \text{EditDist}(ref, hyp^{(i)}) / |ref| \\
\text{BertF1@$k$} &= \max_{i = 1, \ldots, k} \text{BertF1}(ref, hyp^{(i)})
\end{split}
\]
In addition to generation quality, we also measure ranking performance by mean reciprocal rank (MRR@$k$) and success at k (S@$k$) following prior work.
To train the encoder-decoder models, cross entropy loss was used throughout.
All neural models were trained up to 3 epochs (roughly 83k steps) and early-stopped if no further gain on dev WER@3 was observed in the next 10k steps.
At inference time, up to 5 suggestions were generated for each session using beam search (width = 8).
For Seq2Seq+Attn the batchsize was set to 128 and both BART-based methods to 16.
All experiments were conducted on a single NVIDIA T4 GPU.

\begin{table}[t]
    \centering
    \small
    \begin{tabular}{lrrrr}
        \toprule
        & WER & BertF1 & MRR@3 & S@3\\
        \midrule
        Seq2Seq+Attn & 88.3 & 53.1 &  9.8 & 14.1 \\
        MPS          & 49.7 & 72.6 & 33.9 & 47.1 \\
        BART         & 41.5 & 76.1 & 42.5 & 54.6 \\
        MeshBART   & \bf 40.9 & \bf 76.5 & \bf 42.7 & \bf 55.0 \\
        \bottomrule
    \end{tabular}
    \caption{Top-3 test performance. Differences between BART and MeshBART on WER and BertF1 are significant ($p < 10^{-4}$) on the Wilcoxon sign-rank test.}
    \label{t:main_results}
\end{table}

\section{Results}

\paragraph{Quality of Generated Queries.}

We present the effectiveness scores of different generation models in Table~\ref{t:main_results}.
The results show that query generation remains a difficult task: the top-3 beam search output from a standard Seq2Seq with attention baseline achieves on average 0.88 errors per token.
Consistent with prior work \cite{sordoni2015hierarchical}, MPS delivers competitive performance, bolstering its wide adoption in production systems.
The word error rates are pushed down further by a vanilla BART that simply encodes sessions as long sequences, showcasing the superior modeling power of pretrained Transformer networks.
Among these results, MeshBART consistently demonstrates the best effectiveness across all metrics, suggesting that combining critical signals from the given behavioral hypotheses can improve generation quality.

\begin{figure}[t]
    \centering

    \includegraphics[width=0.65\columnwidth]{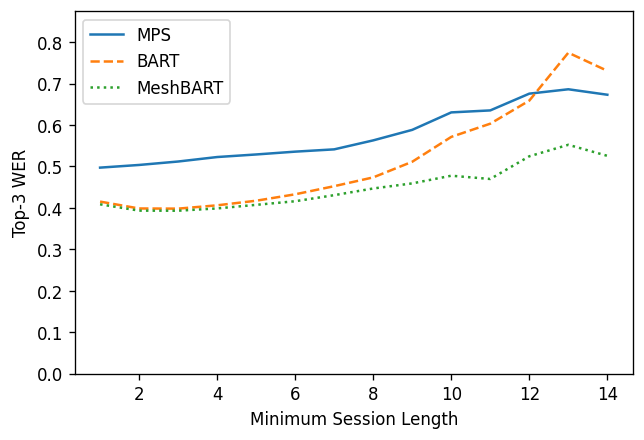}
    \includegraphics[width=0.65\columnwidth]{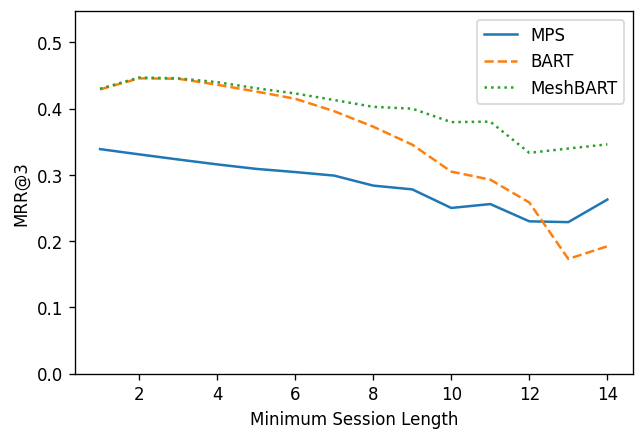}
    \caption{A breakdown of top-3 word error rate and  MRR@3 by minimum session length. Each bucket on x-axis indicates a sub-population of test sessions that contain at least $X$ queries.}
    \label{f:top3-wer-mrr3-session-length}
\end{figure}

We also investigate if generation quality is influenced by other factors of the test population.
Placing a limit on session length, in Figure~\ref{f:top3-wer-mrr3-session-length} 
we find that long sessions result in lower performance across all models.
The diverse and complex intents commonly involved in long sessions make next-query prediction particularly challenging.
Interestingly, WER@3 and MRR@3 appear to respond differently to the increased session length for different methods, e.g. BART can perform worse than MPS on excessively long sessions.
Apart from that, MeshBART remains the most competitive across all buckets, suggesting it being a robust approach for query generation.      

\begin{table*}
\centering
\small
\begin{tabular}{p{3.5cm}p{4cm}p{6cm}}
  \toprule
  Preceding Query ($Q_n$) & Candidates & Generative Suggestions (MeshBART) \\ 
  \midrule
  environmental technology &
  environmental, environment &
  environmental, environmental science, sustainability, environmental scientist \\
  
  part time adobe &
  part time, part time marketing &
  part time marketing, marketing, digital marketing, graphic design \\
  
  aviation security adelaide &
  adelaide airport security &
  aviation security, security, security officer, airport security \\
  \bottomrule
\end{tabular}
\caption{Query generation examples on tail queries, based on test sessions with only one preceding query that the logs fail to produce enough candidates for due to scarcity. Generative models such as MeshBART can produce reasonable suggestions regardless of candidate pool coverage.}
\label{t:examples}
\end{table*}

Further comparisons with non-generative MPS also shed light on the superiority of the proposed approach.
In a W/T/L analysis on top-3 ranking performance, we find that across all test sessions MeshBART has seen 30\% wins, 52\% ties, and 18\% losses to MPS on MRR@3.
Another analysis conditioned on sessions with only one preceding query shows that MeshBART can produce at least one novel suggestion (i.e. queries not seen in the candidate pool) for 39.2\% of the test sessions.
The effect is more pronounced when the preceding query is rare or has a relatively smaller candidate pool.
Examples given in Table~\ref{t:examples} show that the proposed approach can formulate reasonable follow-up queries by generalizing seen query parts.

\begin{figure*}[t]
    \centering
    \begin{subfigure}[b]{0.24\textwidth}
        \includegraphics[width=\textwidth]{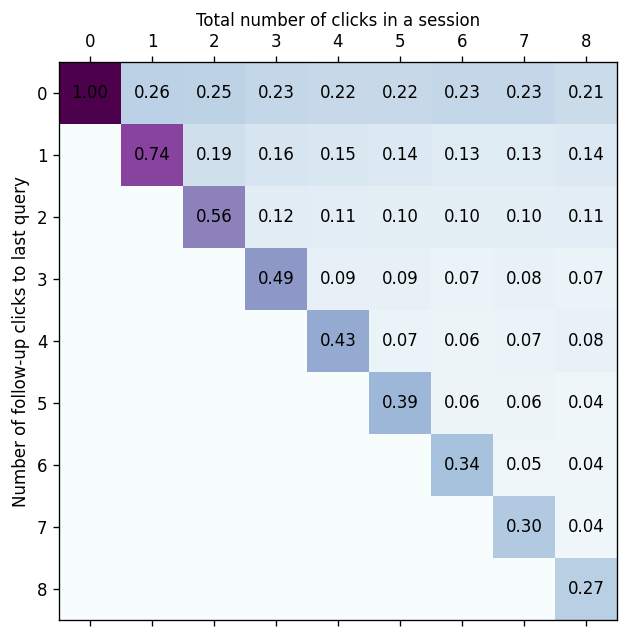}
        \caption{}
        \label{f:corr-heatmap}
    \end{subfigure}
    \begin{subfigure}[b]{0.30\textwidth}
        \includegraphics[width=\textwidth]{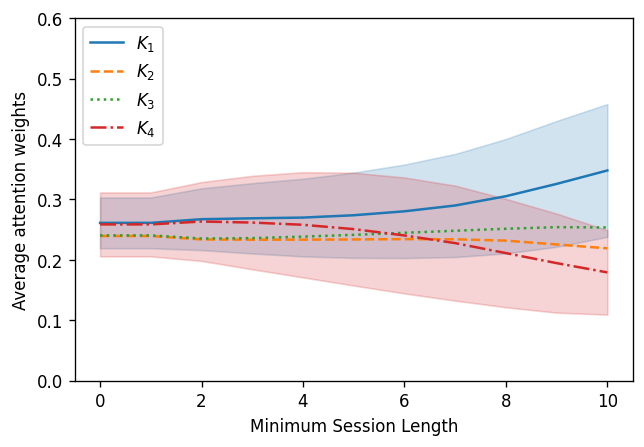}
        \caption{}
        \label{f:attn-session-length}
    \end{subfigure}
    \begin{subfigure}[b]{0.30\textwidth}
        \includegraphics[width=\textwidth]{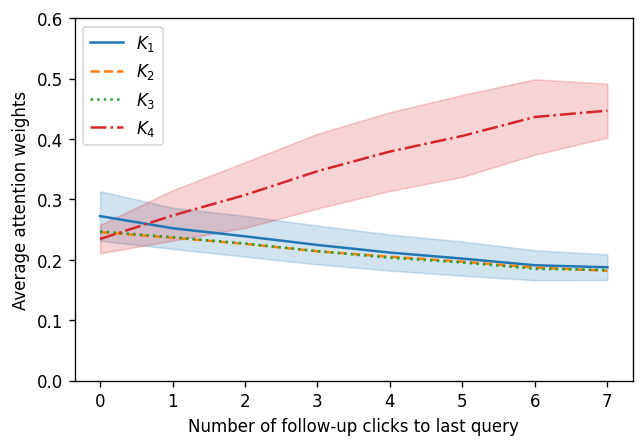}
        \caption{}
        \label{f:attn-num-clicks}
    \end{subfigure}
    \caption{(a) Column-normalized contingency table illustrating clicking behavior. (b)(c) Average attention weights for all four behavioral hypotheses vary across different session type buckets. Shaded areas for $K_1$ and $K_4$ indicate the standard error. (Best viewed in color.)}
    \label{f:top3-wer}
\end{figure*}

\paragraph{Analysis of Behavioral Hypotheses.}

Our design of behavioral hypotheses in Section~\ref{sec:approach} is inspired by the users' interaction patterns.
Figure~\ref{f:corr-heatmap} illustrates the intensity of user clicking with respect to the last query in a session, suggesting that the majority of clicks are predominately centered on the last query irrespective of the total number of clicks.
Job search users are found to be more persistent on articulating an effective query, and from there consume extensively most returned results before disengaging from the search session.
This interesting perk is best reflected by the use of $K_4$ hypothesis in our modeling framework.

To understand the inner workings of the meshed attentions, Figure~\ref{f:attn-session-length} visualizes the actual attention values assigned to all four of the presented hypotheses across different length buckets.
On the one hand, the attention weight associated with $K_1$ (i.e., all preceding queries) are found to positively correlate with the growth of session length. 
Longer search sessions might be due to the user actively exploring the search space, and this increased attribution signals the importance of explicit search intents from the modeling perspective.
On the other hand, $K_4$ tends to receive less attention as the search session grows, indicating that the importance of the most recent interactions become diluted in long, exploratory search journeys.
The value of recent interactions in generative modeling is best illustrated by Figure~\ref{f:attn-num-clicks}, where the attention weight of $K_4$ appears to positively correlate with the intensity of last-round clicking in a search session.
These results suggest that our approach has the flexibility to draw information from different hypotheses in a unified query generation process.

\section{Conclusions}

This paper presents an effective approach for incorporating user-induced interaction patterns as behavioral hypotheses into the query generation process.
Under an encoder-decoder Transformer framework, the proposed tokenwise attentions demonstrate the desirable modeling working by placing emphasis on different behavioral hypotheses at different occasions.
On a domain-specific search benchmark, our model outperforms all reference methods in aggregate and across varying session properties, demonstrating its effectiveness in a robust way.
In future work, we will focus on producing novel continuations of the user's search intent, extending the approach to other domains, and automating the design of behavioral hypotheses.
Qualitative evaluation for open-ended generation is also an interesting topic on the roadmap.

\section*{Acknowledgments}

We would like thank the reviewers for their valuable feedback on evaluation and further analyses.


\clearpage

\bibliography{paper}
\bibliographystyle{acl_natbib}

\clearpage

\appendix

\end{document}